\documentclass{article}
\usepackage{arxiv}
\usepackage[utf8]{inputenc} % allow utf-8 input
\usepackage[T1]{fontenc}    % use 8-bit T1 fonts
\usepackage{hyperref}
\usepackage{ upgreek }% hyperlinks
\usepackage{url}            % simple URL typesetting
\usepackage{booktabs}       % professional-quality tables
\usepackage{amsfonts}       % blackboard math symbols
\usepackage{nicefrac}       % compact symbols for 1/2, etc.
\usepackage{microtype}      % microtypography
\usepackage{lipsum}
\usepackage{float}
\usepackage{graphicx}
\usepackage{multirow}
\usepackage{array}
\usepackage{comment}
\usepackage{longtable}
\usepackage{ragged2e}
\usepackage{subfig}
\usepackage{amsmath}
\usepackage{longtable}
\usepackage{ amssymb }
\title{Case Studies on X-Ray Imaging, MRI and Nuclear Imaging}

\author{
  Shuvra Sarker   \\
  Research and Development Department, Pioneer Alpha,\\
  Dhaka, Bangladesh\\
  \texttt{shvrasarker17@gmail.com} \\
   \And
    Angona Biswas   \\
 Research and Development Department, Pioneer Alpha,\\
  Dhaka, Bangladesh\\
  \texttt{angonabiswas28@gmail.com} \\
   \And
      MD Abdullah Al Nasim \\
   Research and Development Department, Pioneer Alpha,\\
  Dhaka, Bangladesh\\
  \texttt{nasim.abdullah@ieee.org} \\
   \And
         Md Shahin Ali \\
   Department of Biomedical Engineering, Islamic University, Kushtia-7003, Bangladesh\\
  \texttt{shahinbme.iu@gmail.com} \\
   \And
Sai Puppala\\
Department of Computer Science, University of Alabama at Birmingham,Alabama, USA\\
\texttt{spuppala@uab.edu} \\
   \And
 Sajedul Talukder \\
  Department of Computer Science, University of Alabama at Birmingham,Alabama, USA\\
  \texttt{stalukder@uab.edu}}  

\begin{document}
\maketitle

\begin{abstract}
The field of medical imaging is an essential aspect of the medical sciences, involving various forms of radiation to capture images of the internal tissues and organs of the body. These images provide vital information for clinical diagnosis, and in this chapter, we will explore the use of X-ray, MRI, and nuclear imaging in detecting severe illnesses. However, manual evaluation and storage of these images can be a challenging and time-consuming process. To address this issue, artificial intelligence (AI)-based techniques, particularly deep learning (DL), have become increasingly popular for systematic feature extraction and classification from imaging modalities, thereby aiding doctors in making rapid and accurate diagnoses. In this review study, we will focus on how AI-based approaches, particularly the use of Convolutional Neural Networks (CNN), can assist in disease detection through medical imaging technology. CNN is a commonly used approach for image analysis due to its ability to extract features from raw input images, and as such, will be the primary area of discussion in this study. Therefore, we have considered CNN as our discussion area in this study to diagnose ailments using medical imaging technology.

\end{abstract}




\maketitle

\section{Introduction}\label{sec1}

As part of growing advancements in the medical field, Medical imaging played a significant role in terms of diagnosing a variety of ailments. Medical Imaging is the process of viewing and monitoring the interior of the anatomy that aids in the diagnosis and treatment of disorders \cite{MedicalImaging}. Some of the popular techniques including X-ray, MRI, and Nuclear Imaging are used for diagnostic procedures that provide images of the human interior organs and bones. 

The X-ray (Rontgen radiation) is a non-invasive clinical diagnostic technology that yields 2D images of the internal structure of the body by utilizing electromagnetic radiation of extremely high frequency, high energy, and short wavelength \cite{narin2021automatic}. It was invented by Wilhelm Conrad Rontgen, a German scientist, who named it x-radiation due to its unknown type of nature \cite{Xray}. Rontgen discovered that invisible rays from a cathode-ray tube were penetrating through cardboard, causing the fluorescent screen to illuminate. An X-ray beam is carried through the anatomy, with some of the radiation being absorbed or deflected by the interior while the remainder is delivered to a detector (photographic film), resulting in images of the internal structure \cite{CDC}. Less dense tissues appear as gray on the detector as most of the X-rays can pass through the soft tissue, whereas densely packed structures appear as white, e.g., bone. In addition to detecting fractures (broken bones), X-rays can also be utilized to inspect the kidney, bladder, lungs, stomach, and liver \cite{ClevelandClinic}. This helpful feature from X-ray imaging assists doctors in further looking into problems related to bones, joints, soft tissues, and other sections of the body effectively. Moreover, guiding surgeons during coronary angiograms for emergency treatment after a heart attack is another vital role of X-rays.  They can also be used to treat cancer, pneumonia, injuries, breast cancer, and dental issues. Recently, the diagnosis of bacterial pneumonia and the most serious life-threatening COVID-19 disease has relied heavily on X-ray imaging. Several diseases can be diagnosed effectively by employing X-ray images. Among them, several studies have been carried out on COVID-19. Identifying the importance of this topic we have discussed in detail in this chapter about detecting some of the issues related to COVID-19 from x-ray images. Since COVID-19 is a pandemic disease and quick treatment could help in the early identification of bodily issues, Artificial Intelligence (AI) along with this radiological x-ray imaging will assist in diagnosing the disease. The majority of the studies are centered on deep learning, which can diagnose illnesses autonomously through X-ray images.

Another non-invasive medical imaging approach is Magnetic Resonance Imaging (MRI) which yields 3-D detailed, high-quality images of the internal structure of the body from various angles. Although the invention of MRI is done by Raymond Damadian, the Nobel Prize was awarded to Paul Lauterbur and Peter Mansfield for providing a basis for gathering information from high-quality images \cite{GDPR}. MRI scanners are well suited to capture images of the soft tissues of the anatomy. MRI can be employed to examine the blood vessels and heart, internal organs (liver, kidneys, etc.), cancers, and other structures \cite{mayo}. This technology uses radio frequency (RF) and a strong magnetic field instead of damaged electromagnetic radiation. After aligning the protons in the human body a radio frequency is provided which causes the input proton stream to become misaligned. When the RF is shut off, the protons realign, but they continue to send radio signals back, permitting the scanner to gather images of a specific region of the body. MRI is a prominent screening test for the diagnosis of any type of brain or spinal cord disorder. Additionally, AI can evaluate the image for any form of diagnosis in order to gain new discernment about any ailment. MRI has recently become the most effective way to detect brain tumors in the medical field. According to prior research, For automating the identification of this brain tumor using MRI, AI-based approaches yield the most accurate and efficient results.

Nuclear medicine imaging is quite distinct from other conventional imaging modalities. It is a process of tracking radiation released by the body as opposed to radiation produced by other external sources, such as X-rays. It differs from other radiological imaging since it emphasizes the organ's functionality, whereas other methods focus simply on the organ's appearance. It makes use of radioactive substances (radiopharmaceuticals), which are ingested or inhaled into the body, and the detector records images from the radiation released from the radiopharmaceuticals. It focuses on organ tissue, such as lung scans, brain scans, heart scans, bone scans, gallbladder scans, and tumors \cite{nuclear}. This approach may also identify any damage and target the damaged cell for destruction by a tracer to halt the development of that damaged cell.
In this chapter, some of the use cases of medical imaging modalities are briefly explored, with an emphasis on their applicability to the most frequent ailments. This study presents the diagnosis of some particular diseases utilizing medical imaging with the help of AI technologies. This study offers a thorough review of illness diagnosis using X-ray, MRI, and NI with the use of DL techniques studied in numerous research publications. Recently, DNN has achieved remarkable traction in the realm of the medical field for its autodidact capacity to classify disorders. The brief review of the existing methods will assist researchers in obtaining knowledge in this field and conducting a further study on this issue.

An insightful analysis of X-ray, MRI, and NI imaging modalities approaches and their applications are presented in this chapter after reviewing numerous articles. Based on the prior research from several articles, only a few are selected to study the method they used in their article along with the result they achieved, which are presented in brief here covering x-ray, MRI, and Nuclear imaging (NI). 

The below content is organized in the following structure: The related studies on the mentioned topic are presented briefly in section 2. In Section 3, materials and methodology are described. In Section 4, the results and analysis are depicted. Section 5 presents the main findings of this chapter.

\section{Background Study and Related works}
\subsection{Medical Imaging and Essential Study in Medical Science}
\subsubsection{Medical Imaging}
The history of medical imaging dates back to the discovery of X-rays by Wilhelm Conrad Roentgen in 1895. Since then, the field of medical imaging has evolved rapidly with the development of various imaging modalities, such as computed tomography (CT), magnetic resonance imaging (MRI), ultrasound, and positron emission tomography (PET). Here is a brief overview of the major milestones in the history of medical imaging: In 1895, Roentgen discovered X-rays, a form of electromagnetic radiation that can penetrate through the body and produce images of internal structures. X-rays quickly became a popular diagnostic tool for detecting fractures, tumors, and other abnormalities. In the 1950s, ultrasound imaging was developed, which uses high-frequency sound waves to produce images of internal structures. Ultrasound is particularly useful for imaging soft tissues and organs, such as the liver, kidneys, and fetus during pregnancy. In the 1970s, the first CT scanner was developed, which uses X-rays and computer processing to produce detailed cross-sectional images of the body. CT scans are particularly useful for detecting tumors, injuries, and internal bleeding. In the 1980s, MRI was developed, which uses a powerful magnetic field and radio waves to produce detailed images of the body. MRI is particularly useful for imaging soft tissues, such as the brain, spinal cord, and joints. In the 1990s, PET scanning was developed, which uses a radioactive tracer to produce images of the body's metabolic activity. PET scans are particularly useful for detecting cancer, heart disease, and neurological disorders. In conclusion, medical imaging has come a long way since the discovery of X-rays in 1895. The development of new imaging modalities and technologies has revolutionized the field of medicine, enabling doctors to diagnose and treat diseases with greater accuracy and precision. 

\subsubsection{X-Ray in Medical Science} 
X-rays are commonly used to diagnose and monitor bone fractures, tumors, infections, and other abnormalities in the body. X-ray machines produce a controlled amount of radiation that passes through the body and creates an image on a film or digital detector. The resulting image shows the internal structures of the body in shades of gray, with denser structures appearing whiter and less dense structures appearing darker.

X-rays are a type of electromagnetic radiation that has a wavelength shorter than visible light. When X-rays pass through the human body, they are absorbed in different rates by different tissues and structures, depending on their density and composition. This property of X-rays makes them a useful tool for medical imaging, allowing doctors to visualize internal structures and diagnose various medical conditions. 

\subsubsection{MRI in Medical Science} 
MRI technique is particularly useful for imaging soft tissues, such as the brain, spinal cord, and organs, which are difficult to see with other imaging techniques. MRI can detect a wide range of medical conditions, including tumors, infections, injuries, and neurological disorders. It can also be used to monitor the progression of certain diseases and to guide surgical procedures. 

MRI, which stands for Magnetic Resonance Imaging, is a medical imaging technique that uses a strong magnetic field, radio waves, and a computer to create detailed images of the body's internal structures. Unlike X-rays, MRI does not use ionizing radiation and is considered safe for most patients \cite{tonmoy2019brain}.

The procedure of MRI conduction in medical science: the patient lies down on a table that slides into a large tube-shaped machine. The machine creates a powerful magnetic field that aligns the hydrogen atoms in the patient's body. Radio waves are then used to disrupt the alignment of these atoms, causing them to emit signals that are detected by the machine's receiver. These signals are then processed by a computer to produce detailed images of the body's internal structures.

\subsubsection{Nuclear imaging in Medical Science} Nuclear imaging, also known as nuclear medicine imaging, is a medical imaging technique that uses small amounts of radioactive materials, called radiopharmaceuticals, to create images of the body's internal structures and functions. Nuclear imaging is different from other medical imaging techniques because it can show how different parts of the body are functioning rather than just their structure \cite{al2022brain}.

During a nuclear imaging procedure, a small amount of radiopharmaceutical is injected into the patient's body, usually into a vein. The radiopharmaceutical travels through the patient's body and accumulates in the organ or tissue being studied. The patient then lies down on a table and a special camera is used to detect the radiation emitted by the radiopharmaceutical. This radiation is then processed by a computer to create images of the body's internal structures and functions.

Several works have been conducted based on the nuclear medical imaging technique. Firstly, we consider X-ray images for depiction. For better analyzing the x-ray imaging modality, detecting coronavirus from the x-ray image has been considered as an example due to its easily accessible quality. It has been considered the primary diagnostic imaging tool for the Coronavirus. There are several pieces of research on this that permits us to take it as an example. The automated identification of this disease is a source of concern due to its infectious nature and the scarce number of test equipment accessible. As a result, previous research relied on an AI-based technique. Recent advancements in deep learning assist in interpreting X-ray images. Among the prior studies, the most frequent and effective method is utilizing CNN through the X-ray images, as it can differentiate features from other common pneumonia directly from raw data. X-ray images is playing a very crucial role in the ongoing pandemic for detecting any lung abnormalities in a timely manner.

Transfer learning technique was adopted along with CNN to recognize COVID-19 from x-ray images, which achieved a tremendous result \cite{apostolopoulos2020covid}. The key emerging tool is an x-ray, from which the data were obtained in order to detect and categorize whether the patient has COVID-19 or not. Ozturk et al. \cite{ozturk2020automated} proposed a DL-based model named ‘DarkNet’ as a classifier where 17 convolutional layers were used along with different filtering from x-ray images to automatically identify COVID-19 to diagnose for binary class categorization (accuracy of  98.08\%) and multi-class categorization (accuracy of 87.02\%). In \cite{hemdan2020covidx}, COVIDX-NET, a DL classifier, was proposed that was trained by seven distinct structures of DCNN models to interpret patient status with 50 x-ray images. The efficacy of MobileNetV2, one of the classifiers, can be enhanced further for use on smart devices due to its fast computing speed in the clinical field. Narin et al. \cite{narin2021automatic} proposed five pre-trained CNN-based approaches utilizing X-ray images, and they found that the pre-trained ResNet50 model offers the highest accuracy. In that model, three distinct binary classifications were implemented with four classes. Wang et al. \cite{wang2020covid} presented COVID-Net, a deep CNN-based design tool that is a public source and accessible to all for making decisions in diagnosis. A sufficient number of datasets should be necessary to accurately identify and categorize the disease.In \cite{nasim2021prominence} COVID-Net therefore plays a crucial role in providing sufficient and current data for automatic classification. They proposed a method for creating this dataset, which will aid radiologists in correctly interpreting X-ray images. 

MRI employs an intense magnetic field combined with radio frequency to offer comprehensive information about soft tissues, bones, organs, and other body interiors. As it provides a detailed picture of soft tissue with 3-D visualization, the detection of brain tumors has been selected as an example from MRI. High precision is needed for identifying brain tumors since even a small inaccuracy might have fatal consequences. The detailed image from MRI can assist the physiologist in determining the particular disease in a particular area due to its capacity to discern between structure and tissue focused on contrast levels. Several studies have been carried out on the automatic detection of brain tumors. In \cite{kanade2015brain}, the authors proposed an automated brain tumor identification technique based on datasets that have been pre-processed with a diffusion filter. The median filter as well as SWT were employed to de-noise it. Following that, segmentation was conducted, and SWT was utilized to extract features before segmenting and classifying with SVM or PNN. In \cite{saladi2023segmentation}, Saladi et al. proposed a more accurate and precise brain tumor detection method for segmentation purposes that makes use of versatile regularized kernel-based fuzzy c-means (ARKFCM). The removal of noise and segmentation present the most challenging issues for MRI brain tumor diagnosis. Three distinct enhancement approaches, notably HE, CLAHE, and BPDFHE, were employed for preprocessing to remove noise from the MRI image. In \cite{biswas2021brain}, three distinct types of tumors were identified utilizing K-means for segmentation, and features were extracted through DWT and then the feature was reduced by adopting PCA. Finally, ANN was adopted for classification. An efficient training function “Levenberg-Marquardt” was employed in the presented network design. To identify tumors and their location, a quicker R-CNN DL technique presented (RPN) with Region Proposal Network. VGG-16 was adopted as a formation structure in that model \cite{bhanothu2020detection}. The binary classification was described in \cite{tougaccar2020classification}, where VGG-16 and AlexNet were utilized for extracting features. They aimed to first increase the remarkable qualities utilizing the hypercolumn strategy, and then merge the information acquired using both structures. The fittest features are chosen to employ recurrent feature elimination (RFE). Afterward, Support Vector Machine (SVM) was employed for categorization, yielding an overall accuracy of 96.77 \%. In \cite{amin2020brain}, the authors presented a DWT-based CNN model in which four MRI sequences were fused via DWT with DW kernel, yielding a more comprehensive tumor region than a single MRI sequence. Thereafter, a partial differential diffusion filter (PDDF) was employed for noise elimination and then segmentation was done utilizing a global thresholding approach. The segmented images were then fed into the 23-layer CNN architecture for classification purposes.

A retrospective analysis was carried out to diagnose lung cancer utilizing nuclear imaging. In \cite{munley1999multimodality}, SPECT as well as PET(positron emission tomography) were integrated with CT  to diagnose lung cancer with 3D treatment planning. They examined the nuclear imaging technique with CT images to improve the accuracy. PET is frequently used in brain tumor analysis. This technique can distinguish between benign and malignant brain tumors. Information from functional evaluations of each organ is a subject of concern for benign lesions, letting the physician know whether the organ is normally activated or not \cite{moriguchi2013clinical}. In \cite{prieto2011voxel}, the authors evaluated voxel-based criteria in conjunction with dual-time-point 18F-FDG PET for the detection and localization of high-grade malignant tumors and they improved sensitivity compared to standard FDG PET. In \cite{shinoura1997brain}, the authors proposed a technique for identifying brain tumors utilizing carbon-11 choline along with PET and examined its effectiveness. C-11 was administered intravenously to humans and rats with tumors. The tracer's dispersion throughout the brain was assessed. FDG, the best tracer in PET, cannot differentiate between HG tumor and necrosis radiation \cite{wong2002positron}. Furthermore, PET scanning might provoke allergic reactions in individuals, which can be harmful. Additionally, due to their inferior spatial resolution compared to MRI scanning, PET tracers are unable to offer reliable localization of physiological structure \cite{wong2002segmentation}. The authors presented a DL-based approach and examined an approach using conventional CT images and CT images derived from PET. ResNet-18 structure was utilized for pre-processing and transfer learning was then used for classification. They demonstrated that traditional CT provides greater accuracy than CT with PET \cite{park2021performance}. In \cite{papandrianos2021automatic}, detection of coronary artery disease (CAD) was diagnosed utilizing SPECT myocardial perfusion imaging (MPI). A robust CNN structure was employed for feature extraction and further classification to identify MPI images. They approached the RGB-CNN model which can train a model efficiently with a limited dataset. The authors implemented another RGB-CNN model to categorize SPECT data. They proposed three CNN approaches where RGB-CNN was designed from scratch while VGG-16 and DenseNet-121 were implemented employing transfer learning \cite{papandrianos2022deep}. They exaggerated numerous CNN layers to obtain the best combination with 10-fold cross-validation which resulted in high accuracy. They compared their RGB-CNN method with VGG-16 and DenseNet-121 where RGB demonstrates adequate performance with strong findings, VGG-16 functioned effectively, and DenesNet-121 retrieved adequate data.

A lot of research advancements were made in X-ray, MRI, and NI Imaging modalities. We have analyzed a few of the important studies in this chapter. Also, In recent times we also reviewed significant traction towards federated learning. federated learning provides an efficient way of detecting newly discovered variants like COVID-19. Some of the useful findings about federated learning are mentioned in \cite{puppala2022towards} \cite{talukder2022federated} \cite{talukder2022novel}. In ~\cite{hossain2023collaborative} a collaborative federated learning system that enables deep-learning image analysis and classifying diabetic retinopathy without transferring patient data between healthcare organizations has been introduced.  

\section{Materials and Methodology of study}
\subsection{X-ray}
X-ray imaging is the most effective diagnostic tool for identifying serious illnesses such as pneumonia. Recently, this imaging technique has also been utilized to diagnose the highly contagious and severe pandemic caused by the coronavirus. Due to the urgency and rapid spread of this pandemic, early detection is crucial in reducing mortality rates and slowing its transmission. However, diagnosing this disease through X-ray images requires specialized expertise or experienced individuals. Therefore, artificial intelligence (AI) has emerged as the fastest and most efficient approach for automatically detecting the coronavirus in X-ray images. Numerous studies have investigated the use of X-ray scans in diagnosing this pandemic disease, and in this context, we will provide specific examples to highlight the significance of this imaging modality in determining the severity of this condition. Among various AI techniques, this review will primarily focus on the CNN-based deep learning approach due to its superior effectiveness \cite{shah2019brain}.

\subsubsection{Materials:}
The progress impact of DL in medical imaging modalities relies on the text-mined image dataset. A vast variety of existing methods have been studied using X-ray imaging for observing various ailments, particularly lung diseases. The detection of COVID-19 has been taken as an example using X-ray images. A dataset is a primary requisite for diagnosing any disease. The chest X-ray scans are obtained from distinct sources. Cohen JP's dataset (covid-chest x-ray-dataset) \cite{covidchest} is one of the frequently used datasets for amassing X-ray images from patients having COVID-19, which considers images from numerous free sources that are periodically updated. ``ChestX-ray8'' is another database that contains 108,948 anterior-view X-ray images of 32,717 distinct patients with annotated eight illness image descriptors that assist in detecting thoracic disease \cite{wang2017chestx}. 

\begin{figure}[H]
\centering
\includegraphics[height=5.2cm,width=\linewidth]{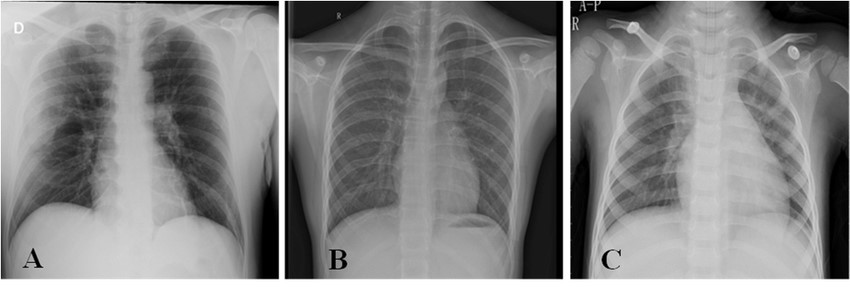}
\caption{Sample images of chest x-ray  (A) COVID-19 case (B) Normal case (C) viral pneumonia case.\cite{wang2020covid}}
\label{xray img.jpg}
\end{figure}

Sample images of the X-ray are shown in Figure 1. This figure depicts three cases of X-ray images. The datasets consist of x-ray images(chest, lungs, bones, and other organs) for recognition and classification purposes.

 \subsubsection{ Methodology:}
 
The COVID-19 virus can cause irreversible lung damage, resulting in pneumonia and perhaps death. Therefore, it is very crucial for doctors to detect corona at a very early stage to reduce mortality. Methods from several studies are presented in this study. The authors used the fuzzy color technique to reform the data classes and stack images that were organized using DL models MobileNetV2 and SqueezeNet \cite{tougaccar2020covid}. The models adopted the Social Mimic optimization (SMO) approach to process the features. Following that, SVM was utilized to aggregate and categorize efficient features. For the first time, Joseph Paul Cohen explored the COVID-19 datasets and publicly shared them on GitHub. The majority of the study for COVID-19 detection is based on this dataset, which aids researchers in their COVID-19 categorization studies. In their research, three different datasets were collected and the images were pre-processed utilizing the Fuzzy technique. The datasets were trained through DL models MobileNetV2 and SqueezeNet and classified using the SVM technique. For image classification, CNN plays a vital role. As a result, the majority of the works are based on CNN-based imaging modalities. 

\subsection{MRI}
\subsubsection{Materials: }

MRI is the most effective and reliable screening technology for detecting brain tumors (BT). It is the best technology that is frequently used as it can diagnose BT at a very early stage. A brain tumor is a source of concern as it is one of the prominent causes of death. As brain tumors do not disseminate to other areas of the body, they can be distinguished from malignant tumors among the numerous types of tumors. Among the three different variants of BT, gliomas form in brain tissues and are considered malignant. Meningiomas develop from the membranes that are benign, while pituitary tumors are masses that rest inside the skull \cite{badvza2020classification}. 
\begin{figure}.
\centering
\includegraphics[height=6.2cm,width=6cm]{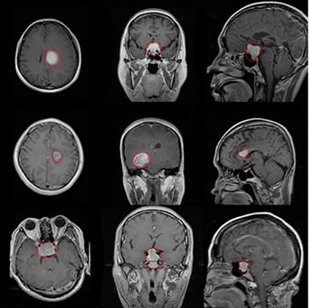}
\caption{Image of MRI brain images \cite{badvza2020classification}}
\label{image.png}
\end{figure}

Figure \ref{image.png} depicts an image of a brain tumor from an MRI. These images are obtained from the datasets for classification purposes or further analysis.

Several public datasets are made available for training the DL models, as the models necessitate an ample amount of data. Three distinct commonly used datasets are utilized in \cite{sahoo2020detection} which were preprocessed and later segmented utilizing Fuzzy-c- means. To categorize BT images, the detection performance of several classification techniques, including discriminant analyzer (DA), K-nearest neighbor (KNN), SVM, decision tree (DT), neural network (NN), and Naive Bays' (NB), had been examined. In \cite{kalaiselvi2020deriving}, the authors developed six distinct CNN models utilizing BraTS 2013 dataset, which was later tested using the WBA dataset. The best result was obtained for the KNN classifier employing the OASIS dataset. Some available datasets from MRI images are depicted in Table \ref{table:1}

\begin{table}[H]
\centering
\caption{Table of some available datasets from MRI images .}
\label{table:1}
\begin{tabular}{c|c} 
 \hline
\textbf{Reference}  & \textbf{Datasets}  \\  
 \hline\hline
 Sahoo et al. \cite{sahoo2020detection},
Kalaiselvi et al. \cite{kalaiselvi2020deriving},
\cite{menze2014multimodal},\cite{papercode} & BRATS  \\ 
\hline
  \cite{oasis} & OASIS   \\
  \hline
 \cite{nitrc} & IBSR  \\\hline
 \cite{kaggle} & Kaggle   \\\hline
 \cite{brainmap} & The Whole Brain Atlas (WBA)  \\ \hline 
 \hline
\end{tabular}

\end{table}

\subsubsection{Methodology: }
The existing approaches employed for brain tumor identification require some common procedures for precise classification. First of all, input images are required for diagnosis and classification which are obtained from the MRI image dataset Nasim et al. The images are then pre-processed through several techniques such as high pass filter, low pass filter, or other filters for eliminating noise. Then data are augmented for training purposes through DL structure and then segmentation is performed. Image processing, thresholding, etc. are done in this segmentation step to find the location of the tumor or other specific tissue. After segmenting the particular area, this is used as input to the DL structure. The preprocessed images can be passed through the DL structure without segmentation. After that, the classification of distinct tumors is performed utilizing several DL approaches such as CNN, ResNet, GoogleNet, and so on, which learn and extract features automatically directly from raw data. CNN is capable of learning patterns from raw images and extracting features automatically. Transfer learning along with DL approaches were used for classification purposes. In \cite{anilkumar2020tumor}, the authors proposed tumor classification where a pre-trained CNN model, VGG-19 was utilized for extracting features, and then KNN was applied for the classification of BT on two autonomous datasets BRATS 20018 as well as CE-MRI. In \cite{gopal2010diagnose}, the authors proposed a more accurate detection procedure for BT utilizing Fuzzy c-means (FCM) clustering with the help of two optimization tools. They suggested two approaches and compared them to determine which was superior. FCM utilizing GA (Genetic Algorithm) method was more time-consuming than the other. As execution is an important parameter to analyze images, their proposed method of PSO (Particle Swarm Optimization) with FCM took less time than the other. Segmentation is a critical step in finding ambiguous regions from complicated medical images. They presented segmentation approaches that made use of the image processing technique FCM together with the different pre-processing tools GA and PSO. The authors proposed a new approach for data augmentation on numerous datasets. Several sets of MRI data were trained to utilize GAN (generative adversarial net) to create MRI-like visuals.
\subsection{Nuclear imaging}
\begin{figure}[ht]
\centering
\includegraphics[height=5.2cm,width=\linewidth]{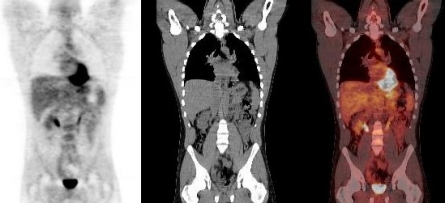}
\caption{Image of nuclear medicine imaging \cite{stanfordhealthcare}}
\label{image3.jpeg}
\end{figure}

Nuclear imaging is an organ-based functional modality that has a significant effect on patient care by offering tools for disease staging. It is an approach for creating images by sensing radiation from distinct locations of the anatomy after a radioactive tracer has been administered to the patient. The benefit of evaluating an organ's function is that it assists clinicians in making plans and developing diagnostic strategies for the area of the body being analyzed. This differs from other imaging modalities in that it employs a tracer with no side effects and minimal radiation. Hence it is safer than the other modalities. Furthermore, it provides some essential information at a preliminary phase that other imaging modalities can’t provide. Following the administration of the tracer and the accumulation of the tracer in the bodily tissue being studied, radiation will be produced and detected by a gamma camera. During a nuclear scan, physicians can analyze and diagnose disorders such as tumors, infections, cysts, etc. by observing the behavior of radioactive substances. The region of the body where the radionuclide accumulates in larger proportions is called hot spot \cite{hopkins}. This modality can detect abnormal renal function, brain, thyroid, and heart function, among other things.
Figure \ref{image3.jpeg} presents an image from nuclear imaging technology.

Nuclear imaging approaches are of two types-SPECT (single-photon emission computed tomography) and PET (positron emission tomography) which yield metabolic and functional details, as opposed to CT and MRI. They are integrated with CT and MRI to disseminate necessary information and permit interconnection between functional and physiological data. Compared to SPECT, PET offers superior contrast and positional accuracy \cite{radiopaedia}. PET makes use of a positron-emitting tracer, while SPECT employs a gamma-emitting tracer. 

SPECT is a 3D imaging modality that enables radiopharmaceutical dispersion as well as provides more clarity, brightness, and spatial information compared to planar nuclear imaging. It employs numerous gamma cameras that spin around the patient to provide greater accuracy and spatial precision. The reconstructed data produces 3D images from cameras \cite{singlephoton}. 

PET is used to assess radioactivity in vivo. It entails injecting a positron-emitting tracer intravenously, allowing for distribution systemically, and afterward monitoring for identification and measurement of radiopharmaceutical dissemination shapes in the anatomy. F-18 fluorodeoxyglucose (FDG) is used as a radiopharmaceutical in the cell. Having high metabolism, tumor cells metabolize FDG which is processed into FDG-6 phosphate [47]. Tumor cells are unable to consume that compound which is later aggregated and concentrated in the tumor cells that fulfill the purpose of detection. Maps of the lung perfusion in 3D are produced using SPECT. Tumor localization is aided by the 3D metabolic pictures produced by PET \cite{hossainbrain}.

A DL-based approach to detect coronary artery disease utilizing SPECT, a nuclear imaging (NI) technique has been taken as an example to show the methodology from the NI technique. Most of the datasets were prepared by the Nuclear Diagnostic Department to diagnose diseases from NI. The dataset for this approach was prepared from a medical center where SPECT images were used to visualize the heart.
In \cite{papandrianos2022deep}, the authors proposed an RGB-CNN model where the SPECT MPI images were first loaded and the images are presented in RGB mode. Following that, the data was prepared through normalization, shuffling, and splitting. The data augmentation was performed and the suggested CNN model was trained. Data augmentation was done due to broadening the limited size of the dataset. Their proposed model was able to identify three classes of heart disease autonomously. They utilized appropriate parameters in their model to correctly classify the cardiac disease automatically and examined using different convolution layers for the best combination from where the 16-32-64-128 combination outperformed.

\section{Result and analysis}
A review of some existing methods has been presented here.

%{footnotesize
%\onecolumn
\begin{table}
\centering
\caption{Performance evaluation of numerous studies for COVID-19 detection from x-ray images:}
\label{table:2}
\begin{tabular} { | p{2cm} | p{2 cm} | p{3cm} | p{1.5 cm} | p{2.5 cm} | p{2 cm}| } 
 \hline
\textbf{Authors} & \textbf{Method} & \textbf{Dataset} & \textbf{Classes} & \textbf{No. of X-ray samples} & \textbf{Accuracy (\%)}  \\   
 \hline\hline
Nayak et al \cite{nayak2021application} & ResNet-34 &	• covid-chestxray-dataset;

•	ChestX-ray8 & 2 &
406

•	Covid19-203

•	Normal-203 & 98.33 \%  \\ 
\hline
  Ucar and Korkmaj \cite{ucar2020covidiagnosis} & Bayes-SqueezeNet & •	covid-chest x-ray-dataset;
  
•	Kaggle chest X-ray pneumonia dataset & 2 & 5949

•	COVID-76

•	Normal-1583

•	Pneumonia -4290
& 98.3\%
   \\
  \hline
 Ozturk et al. \cite{ozturk2020automated} & DarkCovidNet &•covid-chestxray-dataset;
 
•	ChestX-ray8
& 	 Binary class (2) 

&  625

•COVID -125

•Normal-500 & 
98.08\%
\\
\hline
 Toğaçar et al.\cite{tougaccar2020covid} & SqueezeNet, MobileNetV2, SVM, SMO & Stacked dataset of covid-chestxray-dataset and COVID-19-Radioraphy Dataset (kaggle) \cite{radiography} &3 & 458
 
COVID: 295, NORMAL: 65 and 

PNEUMONIA: 98 & 98.25\%

   \\\hline
Hemdan et al.\cite{hemdan2020covidx} &
COVIDX-Net	& Cohen’s covid-chest x-ray-dataset	& 2 &	
50

COVID: 25;

NORMAL: 25	& 90.00  \\\hline
Narin et al. \cite{narin2021automatic} &ResNet-50	& Cohen’s covid-chest x-ray-dataset &2	& 100

COVID: 50; NORMAL: 50	& Dataset-1: 96.1\%

Dataset 2: 99.5\%

Dataset 3: 99.7\%
\\\hline

 Wang et al.\cite{wang2020covid} & COVID-Net	& COVIDX &3	& 13975	& 93.3\%
  \\ \hline 
  Sethy and Behera \cite{sethy2020detection} &SVM, ResNet-50	&Dataset from GitHub, Open-I and Kaggle	&2	&50
  
COVID: 25;
Normal: 25	&95.38\%

 \\\hline
 Farooq, Hafeez and Nasim \cite{farooq2020covid,nasim2021prominence}& COVID-ResNet &	COVIDx	&4	&5941	&96.23\%
\\\hline
\end{tabular}

\end{table}
 From Table  \ref{table:2}, it is observed that numerous studies have been conducted utilizing numerous methods. Among them, the ResNet-34 model achieves 98.33 \% for the detection of COVID-19 disease. X-rays can be utilized to diagnose several diseases, such as heart disease, any type of fracture, lung disease, and so on. The diagnosis of COVID-19 was chosen here as an example utilizing the X-ray images. It is also assumed that the majority of the works used the COVID chest X-ray dataset. 

 Table \ref{table:3} shows that MRI images were utilized to recognize the most common brain tumor. CNN model obtained better results using BRATS datasets.

The nuclear imaging technology was used to diagnose various diseases. Table \ref{table:4} depicts several studies of this imaging technique for coronary disease diagnosis. In that case, the RGB-CNN model performed better. 

In light of the foregoing, it is determined that the majority of imaging modalities employ convolutional neural network (CNN)-based deep learning technology. It is frequently utilized for its best recognition and classification purposes in images. It can classify images as input by discovering patterns on its own. Therefore, it shows remarkable results for image classification. Hence, medical imaging modalities rely on this CNN-based technology for its automatic feature extraction and classification nature.

\begin{table}[H]
    \centering
    \caption{Performance evaluation of numerous studies for brain tumor detection from MRI images:}
\label{table:3}
\begin{tabular}{ | p{2cm}|p{2cm}|p{2.5cm}|p{2.5cm}|p{2.5cm} | } 
 \hline
\textbf{Authors} & \textbf{Purpose }&\textbf{Method} & \textbf{Dataset}  & \textbf{Accuracy (\%)}  \\   
 \hline\hline
Kalaivelsi et al. \cite{kalaiselvi2020deriving} & Detection  along with Classification	&CNN	&BRATS 2013 dataset;The Whole Brain Atlas (WBA)	&96-99 \%  \\\hline
Gopal et al. \cite{gopal2010diagnose}
&Detection, segmentation, and classification 	&Fuzzy c-means algorithm along with  PSO and GA &	

     ---	&GA with FCM:74.6\%
     
PSO with FCM:92.3\%
\\\hline
Togaçar et al. \cite{tougaccar2020classification}
&Binary classification 	&CNN(AlexNet and VGG-16)for feature extraction, RFE for enhancing image and SVM for classification 	&Dataset of Chakrabarty 2019(Kaggle)	&96.77\%

\\\hline
Amin et al. \cite{amin2020brain}
&Binary Classification	&Conventional optimized filter used for pre-processing;
Seed growing technique for segmentation;
Stacked Sparse Auto-encoder(SSAE) for classification	&Multiple BRATS Challenge dataset	&100\% on 2012,

90\% on 2012 synthetic, 

95\% on
2013, 

100\% on Leaderboard 2013,
97\% in 2014,
95\% on 2015

\\\hline
Amin et al. [19]\cite{amin2020brain}
& Classification	&DWT with DW kernel;
CNN for classification 	&BRATS dataset	&Average 97.6 \%
\\\hline
Ghassemi et al. \cite{ghassemi2020deep} & Classify three distinct tumor classes	&GAN with ConvNet
GAN to produce MRI-like visuals; CNN for classification 	&Dataset by cheng et al[55], brain volume MRI images dataset \cite{marcus2010open}
& 93.01\%(introduced split)

95.6\%(random split)
\\\hline
Salma et al. \cite{begum2020combining}
&Detection and Classification 	&optimal wavelet statistical texture
features(OGSA) along with (RNN)	&Images collected from star diagnostic and we resources 	&96.26\%
\\\hline
    \end{tabular}
    \label{tab:my_label}
\end{table}

\begin{table}[H]
    \centering
    \caption{Performance evaluation of numerous studies for coronary artery disease detection from Nuclear imaging technology:}
\label{table:4}
\begin{tabular}{|p{3cm}|p{2.5cm}|p{3.5cm}|p{2.5cm}|} 
 \hline

\textbf{Authors} &\textbf{Method} & \textbf{Purpose }  & \textbf{Accuracy (\%)}  \\   
 \hline\hline
 Papandrianos et al. \cite{papandrianos2022deep} &RGB-CNN	&Three class classification to diagnose CAD	&91.86\%
 \\\hline
 Papandrianos and Papageorgiou \cite{papandrianos2021automatic}
&RGB-CNN	&Two class classification of CAD (ischemic or infarction)	&94\%
\\\hline
Arsanjani et al.\cite{arsanjani2015prediction}
&ML based LogitBoost algorithm	&Early revascularization 	&-\\\hline
Berkaya et al. \cite{berkaya2020classification}
&Two classification techniques employing transfer learning and SVM	&Two class categorization of myocardial anomalies	&94\%
\\\hline\hline
\end{tabular}
\end{table}
%\end{footnotesize}

\section{Conclusion}

This chapter represents several studies on three medical imaging modalities X-ray, MRI, and nuclear imaging for diagnosis purposes. The findings of these studies highlight the superiority of artificial intelligence (AI) approaches in the field of medical science. DL-based models, particularly CNN-based models, are widely used for image classification, eliminating the need for manual feature extraction. This advanced structure can extract features directly from raw data and can be constructed by combining different layers. Although these DL techniques primarily require large datasets, data augmentation can be used to benefit smaller datasets. The images from the dataset undergo pre-processing through various filters, segmentation, and classification, utilizing DL structures with precise features. The automatic recognition and classification abilities of DL techniques from raw data alleviate the workload of doctors and clinical staff.

From the reviewed studies, it is observed that X-ray is the most frequently used technology in the diagnosis of lung diseases. Currently, the ongoing pandemic COVID-19 is detected and classified using X-ray images, which have achieved remarkable success using the DL technique. From the prior studies, it is depicted that the ResNet-50 model achieves a high degree of accuracy in determining COVID-19 disease. According to prior studies, MRI is the most frequently used technology to identify brain tumors (BT) utilizing DL models. The CNN model and SSAE structures outperformed each other in this review study using this imaging technology to identify and classify BT. Another technique Nuclear imaging provides metabolic as well as functional information, which sets it apart from other imaging techniques. Among several studies in this context, the RGB-CNN structure shows better results in diagnosing cardiac diseases. It offers preliminary information that aids doctors in analyzing organ function and radioactive material activity. Using this imaging modality, relatively less work has been conducted. As a result, there are a lot of opportunities for future study employing this modality. 

Despite its valuable contributions, this chapter has some limitations. Firstly, it lacks a comprehensive analysis of the imaging modalities, as well as an explanation of how DL structures work. Instead, it mainly focuses on studies that employed DL models in imaging modalities as part of a diagnostic approach. Nonetheless, the chapter highlights the crucial role that imaging modalities play in identifying illnesses, as they can assist in diagnosing various diseases. To further illustrate this, the study provides a few specific, widely used diagnoses to observe how the analysis of these modalities using DL approaches can be beneficial.

%/usr/share/texlive/texmf-dist/bibtex/bst/spbasic

%%\bibliography{source.bib}
%\bibliographystyle{plain}
%\bibliographystyle{spbasic} 

% \bibliographystyle{plain}
% \bibliography{source}
% \end{document}

%\bibitem{url} National Center for Biotechnology Information, \url{http://www.ncbi.nlm.nih.gov}

\end{document}